\documentclass[10pt,twocolumn,letterpaper]{article}

\usepackage{cvpr}
\usepackage{times}
\usepackage{epsfig}
\usepackage{graphicx}
\usepackage{amsmath}
\usepackage{amssymb}
\usepackage{multirow}

\usepackage[breaklinks=true,bookmarks=false]{hyperref}

\cvprfinalcopy 

\def\cvprPaperID{****} 

\begin{document}

\title{Thinking Outside the Box: Generation of Unconstrained 3D Room Layouts}

\author{Henry Howard-Jenkins,\space\space\space Shuda Li,\space\space\space Victor Prisacariu\\
Active Vision Lab\\
University of Oxford, UK\\
{\tt\small \{henryhj,shuda,victor\}@robots.ox.ac.uk}
}

\maketitle

\begin{abstract}
We propose a method for room layout estimation that does not rely on the typical box approximation or Manhattan world assumption. Instead, we reformulate the geometry inference problem as an instance detection task, which we solve by directly regressing 3D planes using an R-CNN. We then use a variant of probabilistic clustering to combine the 3D planes regressed at each frame in a video sequence, with their respective camera poses, into a single global 3D room layout estimate. Finally, we showcase results which make no assumptions about perpendicular alignment, so can deal effectively with walls in any alignment.

\end{abstract}
\section{Introduction}

3D room layout estimation aims to produce a representation of the enclosing structure of an indoor scene, one consisting of the walls, floor and ceiling which bound the environment, while removing the clutter. These supporting surfaces provide essential information for a variety of computer vision tasks, not limited to augmented reality, indoor navigation and scene reconstruction.

Almost all prior works to produce 2D and 3D room layout estimations introduce heavy constraints on scene geometry, such as the box approximation \cite{izadinia2017im2cad,zhang2014panocontext} or the Manhattan World assumption \cite{flint2011manhattan,xu2017pano2cad}. Where the box approximation is used, rooms are constrained to be rectangular in shape. Manhattan World geometry, on the other hand, slightly eases the constraints of a boxy world; any number of walls are instead allowed to lie along either of two perpendicular directions. Although these assumptions can still lead to highly representative layouts for a great number of indoor environments, it is trivial to produce examples of room geometries which cannot be accurately captured due to the constraints.

\begin{figure*}[t]
    \centering
    \includegraphics[width=\textwidth]{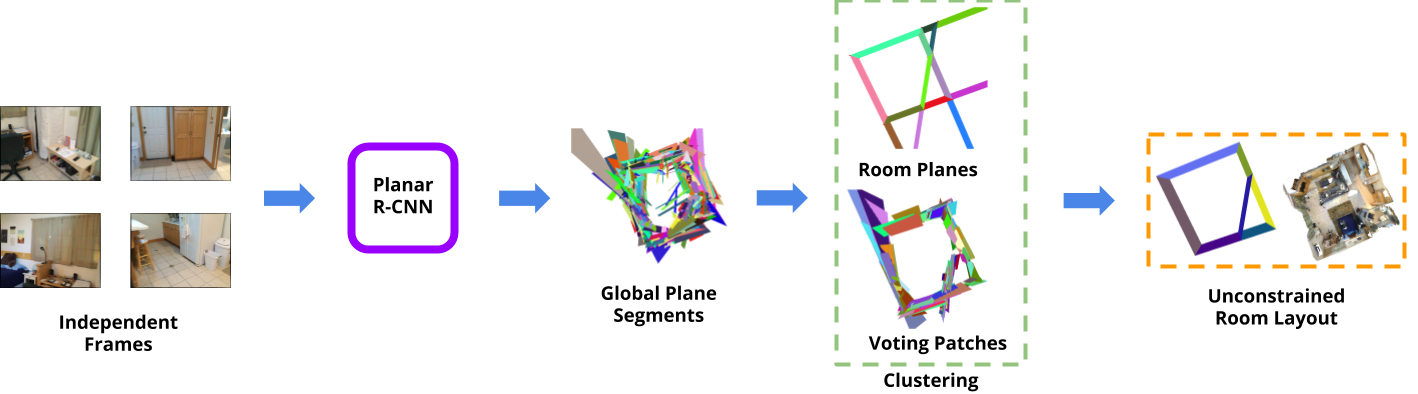}
    \caption{An overview of our method: instantaneous room plane detections are provided by our Planar R-CNN and then combined with their respective camera pose to form a collection of global measurements. Probabilistic clustering is used to infer the number of room planes. These planes are intersected and each segment is considered a candidate wall. Candidates are then accepted into the global room layout based on the support of the measurements, thus capturing the room geometry.}
    \label{fig:tease}
\end{figure*}

The aim of this paper is to relax the constraints on room shape further. In doing so, we are able to represent a wider variety of scene layouts. To accomplish this, we introduce a neural network which is able to detect the presence and the extent, as well as directly regress a representative 3D plane, of room bounding surfaces in single monocular images. By taking the per-image detections accumulated over the frames in a video sequence, along with their respective camera poses, we are able to produce a single global map of plane measurements. From these measurements, we infer the overall geometry of the room without placing constraints on the shape. The scheme is outlined in Figure \ref{fig:tease}.

The main contributions of this paper are: (i) a bounding surface instance detector giving the type and extent of the enclosing walls, floor and ceiling, (ii) direct plane regression for each room plane instance in an RGB image, and (iii) combining the instantaneous measurements to obtain a 3D layout from an RGB image sequence which is not limited to the boxy or Manhattan World constraints.

\section{Related Work}
\label{sec:lit}

Many researchers have sought to produce clutter-free representations of the room layout from monocular images. Early efforts largely had taken the bottom-up approach of extracting image features including colour, texture and edge information to inform vanishing point detection. In these works, a post-processing stage is used to generate, and in turn rank, a large number of room layout hypotheses with either conditional random fields (CRFs) or structured SVMs \cite{gupta2010estimating,hedau2009recovering}. It is possible to recover 3D reconstruction from this approach, but in practice the final layout proves to be heavily dependant on the low level features extracted from the image, which are highly susceptible to noise and clutter. These low level features have also been used to infer 3D spatial layout through depth-ordered planes, by first grouping into lines, quadrilaterals and finally planes.

Mura \textit{et al.} \cite{mura2014automatic} recover accurate floor plans from planes detected in point clouds produced by cluttered 3D input range scans. Even when occluded, planes can be extracted and used to form a set of wall candidates which are projected into 2D.
Rooms present in the floor plan are found by modelling a diffusion process over the the spacial partitions caused by candidate walls. However, this method can only produce 2.5D extruded floor plans and requires a point cloud.

More recently, neural networks have been used to recover spatial layout from RGB images. In \cite{mallya2015learning} fully convolutional networks are trained to replace the earlier hand-crafted features with informative edge maps. The eleven possible representations of the box model room in 2D, as defined in \cite{zhang2015large}, have lead to the problem being formulated as a segmentation problem with classes each representing separate instances of the bounding surfaces: left wall, middle wall, right wall, floor and ceiling as in \cite{dasgupta2016delay}. Furthermore, the eleven representations have been leveraged to tackle layout estimation as an ordered keypoint detection and classification problem as in \cite{lee2017roomnet}, where the room type and the type's respective labelled corners are inferred directly from the image. These 2D representations have been used to produce 3D geometry by informing and ranking room hypotheses \cite{izadinia2017im2cad}.

Panorama images have been leveraged for full scene recovery with box models in \cite{zhang2014panocontext}, and for a Manhattan World approximation in \cite{xu2017pano2cad}. For whole room layout from more traditional camera images, incremental approaches have been used, such as that in \cite{cabral2014piecewise}. Flint \textit{et al.} used a combination of monocular features, with multiple-view and 3D features to infer a Manhattan World representation of the environment \cite{flint2011manhattan}. In \cite{song2017im2pano3d}, a RGB-D panorama is split and half of the scene is taken as input with the aim of producing reasonable, cluttered room layout estimation for the unseen portion of the panorama. 

Alternatively, maps of the environment can be produced iteratively using a top down approach, such as the mapping and planning network architecture detailed in \cite{gupta2017cognitive}, where an egocentric map forms a top down 2D representation of the environment. Once again the representation of the scenes is constrained to 2D and the mapping component aims to recover the navigable space in a scene, rather than the clutter free representation which we seek.

The closest work to this paper is the direct plane regression from a single RGB image performed by Liu \textit{et al.} in \cite{liu2018planenet}. They are able to produce clutter-free room representations from single images. However, their method relies on visible planar regions and therefore room-layouts cannot be produced where a wall, floor or ceiling is completely occluded.

In this paper, rather than focusing on producing a self-complete 2D layout segmentation, we aim to recover the underlying 3D room planes in each image.  Further, wall, floor and ceiling would generally be considered uncountable \textit{stuff}, rather than countable \textit{things} for which object detectors are most commonly used. However, there is work showing that \textit{stuff} classes provide useful context for \textit{thing} detection \cite{brahmbhatt2017stuffnet,mottaghi2014role,rabinovich2007objects,shi2017weakly}. We take this further by treating these architectural planes as \textit{things}, which we detect using an instance detector. These planes are then used to compile a singular representation of the scene which does not rely on any assumption about the geometric structure of the world other than that rooms are planar.

\section{Planar R-CNN}
To address the challenges presented by a non-boxy room layout, we approach the task of finding a room's supporting surfaces as an instance detection problem. In doing so, our network detects an arbitrary number of surfaces and outputs their extents in the form of a bounding-box. Furthermore, by removing the Manhattan World assumption, the walls, floor and ceiling can have any orientation and are no longer just aligned with one of only two dominant directions or the ground plane. Therefore, our network also outputs arbitrary alignment for each of the detection by directly regressing the 3D plane equation of each surface.

We choose to approach the task by directly targeting abstract room-planes rather than using a method which relies on visible patches of wall, floor and ceiling. The advantage gained is two-fold: (i) we are able to infer room planes that are completely or mostly occluded by furniture; (ii) we are able to disregard large planar surfaces that are irrelevant to room structure. On the first point, furniture can often block the wall behind, thus reducing the ability to capture the plane from visible pixels. Considering the second point, the occluding object itself may be the dominant plane within a region, such as the surface of a table.

In all, we are able to predict an arbitrary number of room-bounding planes, as well as their directions and extents, providing all the information required to construct an unconstrained layout.

\subsection{Network Architecture}

\begin{figure*}
    \centering
    \includegraphics[width=0.85\textwidth]{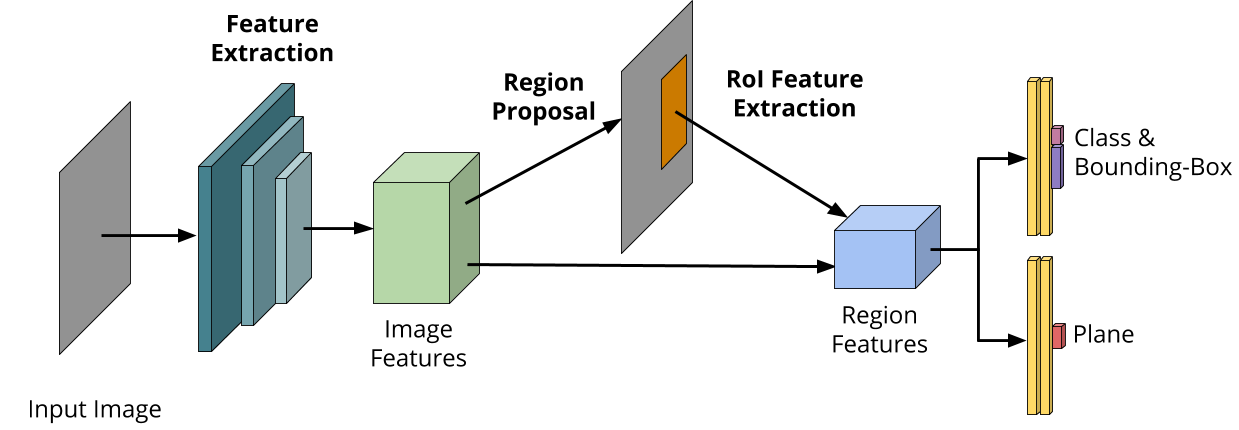}
    \caption{An overview of the complete network from an RGB input image to the outputted classes, bounding-boxes and plane equations. The single dimensional columns in the network heads represent fully connected layers. Yellow columns have an output size of 4096, the plane has output size 4, and the class and bounding-box columns have output size equal to and four times the number of classes, respectively}
    \label{fig:whole_net}
\end{figure*}

We design a network architecture capable of detecting an arbitrary number of walls, floors and ceilings, as well as their extents, positions and alignments. Inspired by Mask R-CNN \cite{he2017mask}, where Faster R-CNN \cite{ren2015faster} is augmented by an additional network head which outputs a mask for each detection, we take a similar approach by adding an additional head to regress a 3D plane equation for each region of interest (RoI). From an input of a single RGB image, the network outputs a number of detections. Each detection has an associated class (wall or floor/ceiling), bounding-box and plane equation.

For clarity, we use the same nomenclature as in \cite{he2017mask} when referring to: (i) the \textit{backbone} which is a convolutional architecture to extract features over the entire input image, and (ii) a network \textit{head} as an architecture applied separately to each set of RoI features, for example the Fast R-CNN head which predicts class and regresses a bounding-box.

\subsubsection{Instance Detection.}
For detection, assigning of classes and determining bounding-boxes for instances of visible room planes, we use an implementation of the Faster R-CNN detector. Faster R-CNN is composed of two stages: the Region Proposal Network (RPN) and a second stage which is essentially Fast R-CNN \cite{girshick2015fast}. 

\subsubsection{Direct Plane Regression.}
In a similar fashion to Mask R-CNN, we introduce an additional network head to the second stage of Faster R-CNN. This network head directly regresses the 3D plane equation of the candidate in parallel to the classification and bounding-box regression. The plane regression head takes the same RoI features as the classification and bounding-box regression head. It outputs a 4-vector which represents the plane coefficients where a 3D point, $\mathbf{x}$, lies on plane $\mathbf{p} = [\mathbf{n}\  d]^T$ if $\mathbf{n} \cdot \mathbf{x} + d = 0$. The first 3 components of the 4-vector, which represent the normal of the plane, are normalised to unit length. 
The 4th plane coefficient, $d$, takes on a more direct physical meaning as the shortest distance between the plane and the camera origin.

The network head itself is comprised of two fully connected layers with 4096 output channels each, in addition to another fully connected layer to take the dimensions down from 4096 to the required 4 representing the plane equation coefficients, as detailed in Figure \ref{fig:whole_net}. The plane head is trained using two losses, $L_{norm}$ and $L_d$, which represent the loss contributed by the normal of the plane and the depth coefficient, respectively. $L_{norm}$ is defined as the negative of the cosine similarity between predicted and ground truth normals, and $L_d$ is the mean-squared-error of the depth coefficients. Both are computed as follows:

\begin{equation}
    L_{norm} = - \dfrac{1}{m} \sum_m \mathbf{n}_{p} \cdot \mathbf{n}_{gt} \quad and \quad
    L_{d} = \dfrac{1}{m} \sum_m (d_{p} - d_{gt})^2
\end{equation}

\noindent where $m$ is the number of predicted planes, $\mathbf{p}_{p} = [\mathbf{n}_{p} \  d_{p}]^T$, and their corresponding ground truth planes, $\mathbf{p}_{gt} = [\mathbf{n}_{gt} \  d_{gt}]^T$.

\subsection{Implementation}
We use an implementation of Faster R-CNN with a VGG-16 \cite{simonyan2014very} as its backbone. The final pooling and all the fully connected layers from the VGG-16 network are removed. In addition to RoIPool from the default Faster R-CNN implementation, we also implement RoIAlign \cite{he2017mask}. While both methods achieve the same general goal of gathering the features relevant to a proposed region, RoIAlign better preserves the spatial information of the region's features, which may lead to improved performance for the spatially sensitive task of plane regression.

The loss for each sampled RoI is defined as $L = L_{cls} + L_{box} + L_{norm} + kL_d$ where $L_{norm}$ and $L_d$ are as above. The classification loss, $L_{cls}$, and the bounding-box loss are as in \cite{girshick2015fast}. We found it was necessary to introduce a weight constant, $k$, for $L_d$ to keep the network stable while training. Setting $k = 0.05$ provided good stability, while ensuring that the depth coefficient still trains.

\section{3D Room Layout Generation}
\label{sec:3d}
The network described in the previous section outputs a number of plane segments representing the room geometry. However, since we do not assume that all rooms are convex, the geometric information predicted from a single image is not complete in itself. For example, there may be entire walls occluded by others. Therefore, we treat each set of plane segment predictions as local measurements from a global room layout.

We define the global room layout as a set of 3D plane segments, each produced by the intersection of underlying room planes. The first task in determining the room layout is to obtain these room planes. We transform the regressed planes and corresponding bounding box measurements from each frame into the global coordinate system. This represents an amalgamation of noisy plane segment measurements each belonging to one of an unknown number of bounding planes. 

We adopt a Bayesian approach for a Gaussian mixture model, and infer the posterior distributions of the room's characteristic planes through EM training. The number of room planes is determined by placing a threshold on the weights in the mixture model.

\subsection{Spatial Voting}
Once we have obtained the room planes, we intersect them all to form a collection of candidate plane segments to add to the room layout. In the convex case, these intersections would provide an immediate layout, however, since we allow non-convexity we must evaluate whether a candidate segment should be added to the room set. To do this, we make use of the regressed bounding-boxes as extents for the predicted planes. Since each measured plane is spatially constrained, it is used to vote for a candidate segment's existence in the room geometry.

We assign every evidence segment to a mixture component distribution. For each of these distributions, we then define a subset of $n$ voter segments by ranking the likelihood of the segment having come from the cluster distribution. This ensures that the measurements lie on the most similar planes to the candidate segments for which they will vote. Support for a wall segment is determined through a robust inlier weighting scheme, such as in \cite{chetverikov2005robust}. A voting plane patch is considered an inlier for a candidate wall by thresholding the proportion of the voter that overlaps the candidate, $i_{vc}$, and vice versa, $i_{cv}$. 

Every candidate is assigned an energy, $\eta = 0$, which is updated on each of the voting segments belonging to the same plane cluster. If a voting patch is considered an inlier, the candidate's energy is increased by $(1 - i_{vc})$, and the number of inliers is incremented. Once all voting patches have been tested, the ratio of inliers to the total voting on the candidate, $r_c$ is computed and the candidate's energy is divided by $r_c^a$, where the exponent, $a$, is a tuneable parameter. To prevent cases where candidate patches are accepted with only a small total number of voter patches, we threshold total energy. The final result of the voting stage is to reject plane segments without sufficient evidence. In all, leaving only the walls that belong to the global room layout.

\section{Experimental Evaluation}

We train and evaluate our network's performance, both quantitative and qualitatively, capturing room geometry from an image in the SUN RGB-D dataset. We compare the regressed planes to their corresponding ground truth instances. In addition we plot the captured planes and their extents with an overlaid point cloud to demonstrate the network's ability to recover bounding planes.

Further, we evaluate our room layout estimation on single frames by testing on the NYUDv2 303 dataset \cite{zhang2013estimating}. We demonstrate the ability of our method to capture useful room layout information by using the output of our network to produce a prediction at the 2D room segmentation for an indoor image.

Finally, we prove the informative nature by amalgamating the independent frame-by-frame predictions on ScanNet image sequences into the global coordinate system and using these measurements to infer a whole room layout. This demonstrates the robustness of the enclosing surface detection and plane regression, since the network was never trained on any ScanNet sequence.

\subsection{Training using SUN RGB-D}
To train our network, we use the SUN RGB-D dataset \cite{song2015sun}. The dataset provides room corner coordinates on the $x-z$ plane, as well as the upper and lower $y$ height of the room, all defined in the world coordinates. This information allows us to build a 2.5D extruded floor plan for the room for each pair of RGB and Depth images in the dataset. We transform the room model from the world coordinate system to one aligned with the camera. From this aligned floor plan, we compute the plane equation of each section of wall, floor and ceiling. The floor plan is then projected into the camera, providing us instance-level segmentation, as well as corresponding plane equations, for every image.

Since we do not test on the benchmarks provided as part of the SUN RGB-D dataset, we use our own image split comprising of 9335 training, 500 validation and 500 testing images. It was found that some of the images in the dataset do not appear to have corresponding room layout information. In addition, for some images, it was found that the picture was taken outside of the annotated room. This tended to occur when pictures were captured in a department store and the room corners were defined as the limits of the display cubicle. We removed these images from the dataset for training, validation and testing. This resulting in the final split containing 7677 training, 441 validation and 439 test images. It is worth noting that the depth images are not used in either training or prediction: instead they are only used to help visualise results.

Bounding surface instances are divided into two classes, one representing walls and the other representing floors or ceilings. This is a result of trying to keep the plane segments as generic as possible, but still maintaining the differences in their representation in the dataset. Because the rooms in the dataset are 2.5D rather than complete 3D floorplans, walls always have normals in the world $x-z$ plane, whereas floors and ceilings must have normals in the world $y$ direction.

\subsubsection{Training}
The models were trained on the dataset for 25 epochs in total. To initialise the feature extractor and classifier in the class and location head, we use a VGG-16 model pretrained on ImageNet \cite{krizhevsky2012imagenet}. Other fully connected layers in both network heads are initialised from a zero mean normal distribution with variance of 0.1, 0.01, 0.001  for the final plane regressor layer, the final classifier layer and the two hidden layers in the plane head, and the bouding-box regressor respectively.
We use a batch size of 1 with stochastic gradient descent, 0.5 dropout rate, 0.0005 weight decay, 0.9 momentum. The learning rate is initially set to 0.001 and decreased to 0.0001 after 20 epochs. This training schedule takes around 12 hours on a single Nvidia GTX 1080Ti.

\subsection{2D Room Layout Estimation with NYUDv2 303}
We explore the generation of 2D layouts using our network, and thus demonstrate its ability to capture room information. As discussed in Section \ref{sec:3d}, our method does not require the convex box approximation. Since we do not impose any such constraints, there is no set number of possible room layouts, or wall types. We do not seek to classify left or right wall, instead we just detect the presence of a wall, floor or ceiling.

To produce predictions, we first back-project each pixel within a detection's bounding box onto its predicted plane, forming a plane patch in 3D. We then form a clutter-free 2D room segmentation by project all of an image's plane patches back into the image. Our reprojection into 2D
provides insight into the usefulness of the plane detections for capturing room layout without having to place constraints on the layout, as well as providing comparison against other state-of-the-art solutions to the layout estimation problem.

Since this dataset takes a subsample of the NYUDv2 dataset \cite{silberman2012indoor}, which is included in the SUN RGB-D dataset \cite{song2015sun}, we retrained our network ensuring that none of the 101 test images appeared in our training or validation dataset.

\subsection{Whole Room Layout Estimation using ScanNet}
In order to demonstrate not only the informative nature of the measurements produced by our network, but also its versatility, we produce 3D room layouts on the ScanNet dataset \cite{dai2017scannet} having never trained on it and without the network using intrinsic data for each sequence. For this task we assume that camera poses are known, as in this case they are implemented using BundleFusion \cite{dai2017bundlefusion}.

\begin{figure*}[t]
	\centering
	\includegraphics[width=\textwidth]{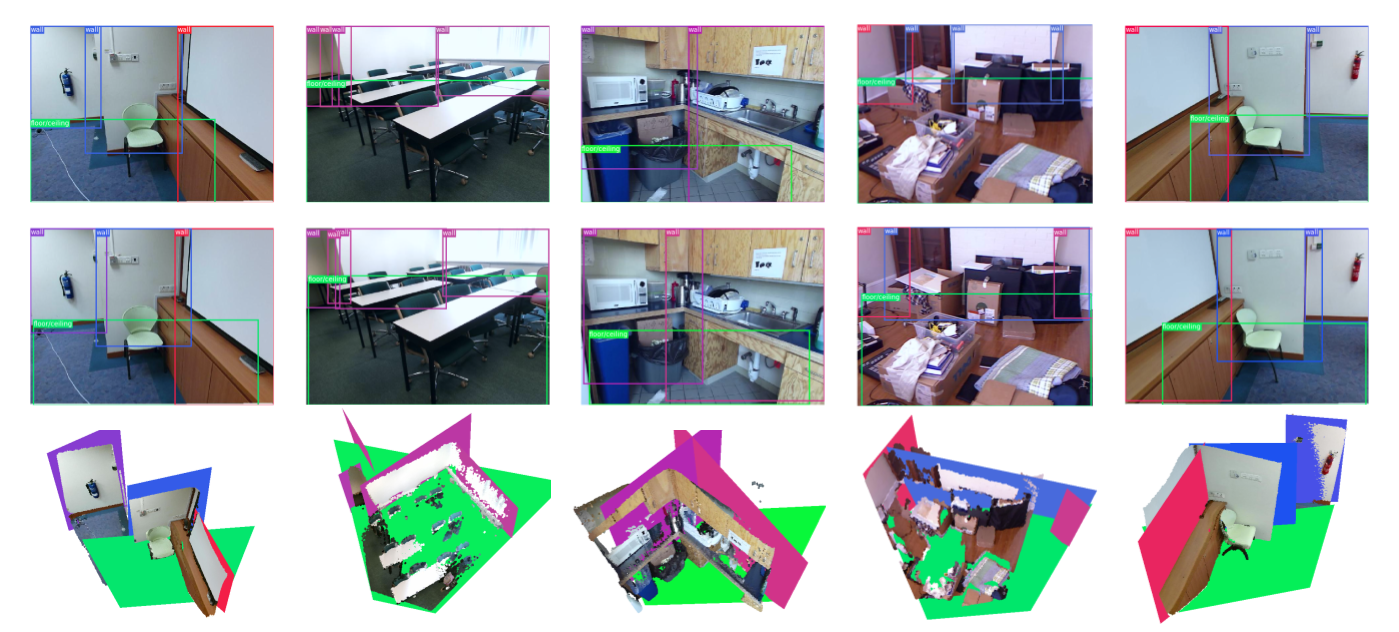}
	\caption{The first and second rows show the image overlaid with the ground truth and network detections, respectively. The last row shows the predicted bounding-boxes projected onto their matching regressed planes to produce the informative plane segments. For additional 3D context, a point cloud from a depth image matching the input RGB image is displayed with the plane segments}
	\label{fig:suc_patches}
\end{figure*}

\subsubsection{Implementation}

We start by collecting all the plane detections in a ScanNet image sequence. Since bounding-boxes do not provide tight constraint on the extent of a plane, especially if the plane is nearly perpendicular to the image plane. Therefore, when collecting instances from images, we discard those surfaces which are less than $30^\circ$ from perpendicular to the image plane.

We found that our spatial voting method for determining which candidate wall planes was sensitive to changes in  set parameters. We believe that this is due, at least in part, to the lack of any training on the ScanNet dataset. Planar R-CNN does not explicitly infer the camera's intrinsics. Therefore, for spatial information recovery at full accuracy, the network should be fine-tuned on new datasets to allow the relation between image coordinates and features with the surrounding 3D environment to be re-assessed. To account for this, we adjusted parameters on a per-scene basis when constructing room layouts. Generally, the parameters were set in the vicinity of: the minimum weight for an accepted plane posterior as 0.05, the 100 most likely measurements from each wall's posterior distribution allowed to vote for existence of the plane's candidate segments, the energy update exponent set to 2, and finally the mutual overlap threshold as more than 70\% of the voter's area and more than 20\% of the candidate's area for the voter to be labelled an inlier for the candidate.

Almost all of the ScanNet sequences have only brief appearances of the ceiling. To provide an upper bound on the wall segments, we instead tested the normal of the dominant posterior from planes labelled as the floor or ceiling class. If the direction of the normal was in the positive $z$ direction, we labelled the plane floor and created a ceiling plane 2\,m above and with a normal in the other direction. If a downward facing dominant plane was detected, it was labelled ceiling and a floor plane was created 2\,m below it.

\subsection{Results}

\setlength{\tabcolsep}{4pt}
\begin{table*}[t]
\centering
\caption{Evaluation of the normal regression on our test set from the SUN RGB-D dataset.}
\label{tab:normals}
\begin{tabular}{c|ccc|ccc}
\multirow{2}{*}{\textbf{Normals}} & \multicolumn{3}{c|}{Error} & \multicolumn{3}{c}{Accuracy ($\alpha < \theta$)} \\
 & mean & median & rms & $11.25^\circ$ & $22.5^\circ$ & $30^\circ$ \\ \hline
FCRN + Room & 33.5 & 20.0 & 44.8 & 36.4 & 52.7 & 59.7 \\
FCRN + Fine & 28.6 & 17.7 & 38.5 & 38.7 & 56.7 & 64.7 \\ \hline
PlaneNet & \textbf{8.35} & \textbf{6.31} & \textbf{13.4} & \textbf{82.3} & \textbf{96.9} & \textbf{97.8} \\\hline
Ours (RoIAlign) & 11.3 & 7.44 & 17.0 & 70.3 & 89.5 & 93.7 \\
Ours (RoIPool) & 9.74 & 6.67 & 15.3 & 76.4 & 92.9 & 94.9

\end{tabular}
\end{table*}
\setlength{\tabcolsep}{1.4pt}

\setlength{\tabcolsep}{4pt}
\begin{table*}[t]
\centering
\caption{Evaluation of the location of the regressed plane compared to its corresponding ground truth plane on our test set from the SUN RGB-D dataset.}
\label{tab:distance}
\begin{tabular}{c|cc|ccc}
\multirow{2}{*}{\textbf{\begin{tabular}[c]{@{}c@{}}Plane\\ Location\end{tabular}}} & \multicolumn{2}{c|}{Error} & \multicolumn{3}{c}{Accuracy} \\
 & \multicolumn{1}{c}{mean} & \multicolumn{1}{c|}{median} & $\delta < 0.2m$ & $\delta < 0.5m$ & $\delta < 1m$ \\ \hline
FCRN + Room & 1.83 & 0.357 & 33.3 & 60.7 & 85.7 \\
FCRN + Fine & 1.01 & 0.242 & 43.7 & 74.0 & 89.2 \\ \hline
PlaneNet & 1.79 & 0.275 & 40.0 & 67.8 & 84.3 \\\hline
Ours (RoIAlign) & 1.18 & 0.269 & 42.0 & 67.4 & 84.1 \\
Ours (RoIPool) & \textbf{0.503} & \textbf{0.217} & \textbf{47.1} & \textbf{75.5} & \textbf{90.7}
\end{tabular}
\end{table*}
\setlength{\tabcolsep}{1.4pt}

Since, to our own knowledge, this network is the first to directly regress abstract room-planes, we consider its main tasks when evaluating its performance. We define these tasks as accurate detection and plane regression. The detection task is evaluated using the average precision for each of the two classes, and the mean average precision. For plane regression, the network must be able to discern the alignment of the plane, \textit{i.e.} the normal of the surface, and the depth coefficient in order to fix the plane in 3D space. We propose a two stage evaluation of the plane regression; first, the quality of the normals, secondly, the spatial accuracy of the plane.

\begin{figure*}[t]
	\centering
	\includegraphics[width=0.95\textwidth]{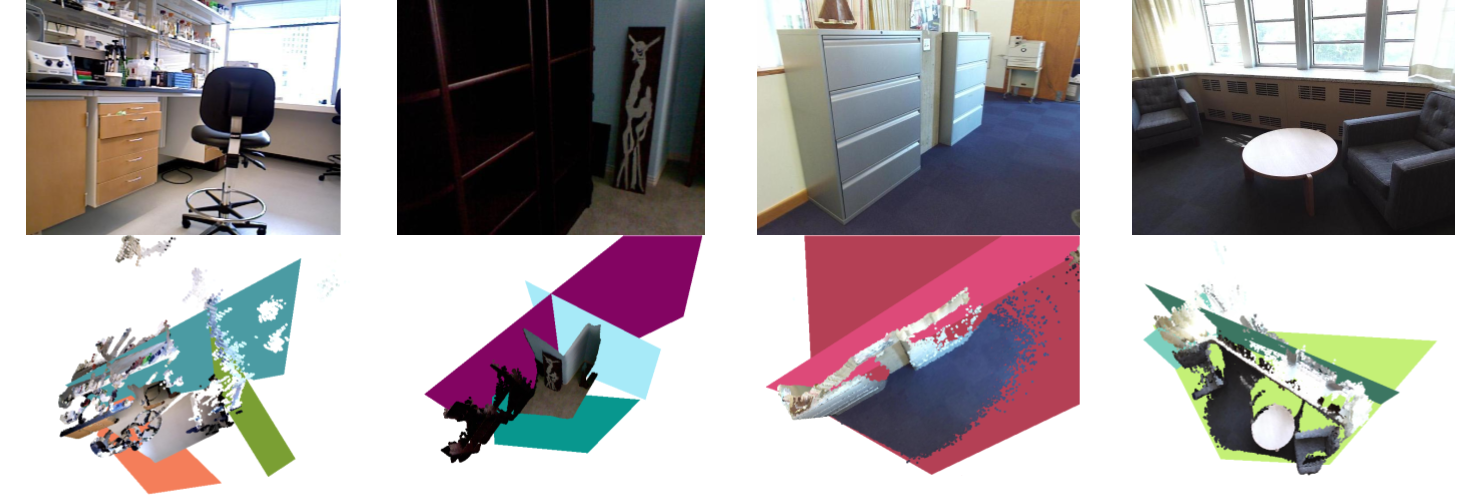}
	\caption{Examples demonstrating the discussed advantages: The first column shows a plane being accurately regressed in a cluttered environment; The second and third columns show examples walls are regressed behind shelves and cabinets; In the forth column a wall is correctly estimated behind a chair and a non-planar curtain.}
	\label{fig:examples}
	\setlength{\textfloatsep}{9pt}
\end{figure*}

\subsubsection{Instantaneous Detections on SUN RGB-D} We assess the quality of the plane normal predictions from the network in the same manner as surface normal estimation from monocular images, as in \cite{chen2017surface,eigen2015predicting,wang2015designing}. However, we compare the normal of a detected instance to its corresponding ground truth plane, instead of at each pixel. The angular disparity each pair, $\alpha$, is computed. We then compute the mean, median and root-mean-square of $\alpha$, as well as listing the percentage where the error falls below the thresholds $11.25^\circ$, $22.5^\circ$ and $30^\circ$. These results are shown in Table \ref{tab:normals}.

To determine the accuracy of the regressed plane in terms of spacial location, we back-project all the pixels within the predicted bounding-box and on to the predicted plane. For each of the back-projected points, we compute a point-to-plane distance between the point and the corresponding ground truth plane. The mean of these distances are averaged for each instance detection in the image, providing a single distance score, $\delta$, for every predicted bounding plane. In Table \ref{tab:distance} we show the mean and median of these instance distances, as well as the percentage which fall within the thresholds 0.2, 0.5 and 1 metres.

We compare our normal and plane location results to a baseline visible pixel-based method. This method produces a pixel-wise depth estimation using the FCRN method of \cite{laina2016deeper}. The depth map is then masked by a ground truth segmentation of the image and back-projected into 3D space. We use RANSAC to fit a plane to the 3D points corresponding to each ground truth room surface. \textit{FCRN + Room} in Tables \ref{tab:normals} and \ref{tab:distance} uses the ground truth room-layout segmentation to mask the estimated depth map, meaning that all depth points may influence the plane fitting. However, \textit{FCRN + Fine} leverages the full segmentation and only points labelled wall, floor, ceiling or window are included in the plane fitting.

We also compare against PlaneNet, using the method described in \cite{liu2018planenet} for inferring room layout. This takes advantage of the consistent ordering of plane inference of the PlaneNet network, where floors are predicted at a specific index etc. We fine-tuned PlaneNet on SUN RGB-D using the depth-only method mentioned in their paper, this did not outperform their pretrained model.


The results against the baseline in Tables \ref{tab:normals} and \ref{tab:distance} demonstrate the quality of the planes regressed using our method and further illustrate the advantages of targeting the abstract room surfaces, rather than relying on visible pixel patches. In addition, it is worth noting that the baseline makes use of ground truth segmentation. For a true comparison, segmentation predictions should be used, further reducing the quality of the fitted planes. Even with the advantage of ground truth segmentation, the baseline is out performed by our method.

While PlaneNet slightly outperforms our method for the accuracy of the regressed normals, our method predicts room planes that are closer to the groundtruth. We would suggest that the high normal accuracy shown by PlaneNet could be explained by planar objects, such as cupboards and closets, which are generally arranged to be parallel to the walls that they occlude. Further, PlaneNet tends to miss planes which are not near the camera. It is worth noting that the architectural planes which are missed by PlaneNet do not negatively impact the results shown in Tables \ref{tab:normals} and \ref{tab:distance}. Whereas, our method performed well as a room-plane detector, with the RoIAlign and RoIPool methods each achieving a mAP of 67.9 and 69.6, respectively.


Example detections from our network, as well as their respective planes from single images, are pictured in Figure \ref{fig:suc_patches}. Predicted planes are limited in extent in this figure by the predicted bounding-box, in order to produce a number of plane segments. In each case, a colourised point cloud produced from the paired depth image is shown to provide context for the predictions. Also included are examples which break the box and Manhattan world assumptions. Further, in Figure \ref{fig:examples} we show example detections demonstrating the advantages of targeting abstract room-planes over plane-fitting to visible wall, floor or ceiling patches. These examples include walls being correctly detected and their plane being accurately regressed when it is entirely or mostly occluded. 

\begin{table}[t]
\centering
\caption{The performance of our Planar R-CNN for predicting 2D room layouts. Evaluated on the NYUDv2-303 dataset \cite{zhang2013estimating}. Results for other methods are as stated in \cite{liu2018planenet}}
\label{tab:layouts}
\begin{tabular}{c|c|c}
\textbf{2D Layouts} & Input & Pixel Error \\ \hline
Zhang \textit{et al.} \cite{zhang2013estimating} & RGB+D & \textit{8.04\%} \\ \hline
Zhang \textit{et al.} \cite{zhang2013estimating} & RGB & 13.94\% \\
Schwing \textit{et al.} \cite{schwing2012efficient} & RGB & 13.66\% \\
RoomNet & RGB & 12.96\% \\
PlaneNet & RGB & 12.64\%  \\ \hline
Ours & RGB & \textbf{12.19\%}
\end{tabular}
\end{table}

\begin{figure*}[t]
	\centering
	\includegraphics[width=0.75\textwidth]{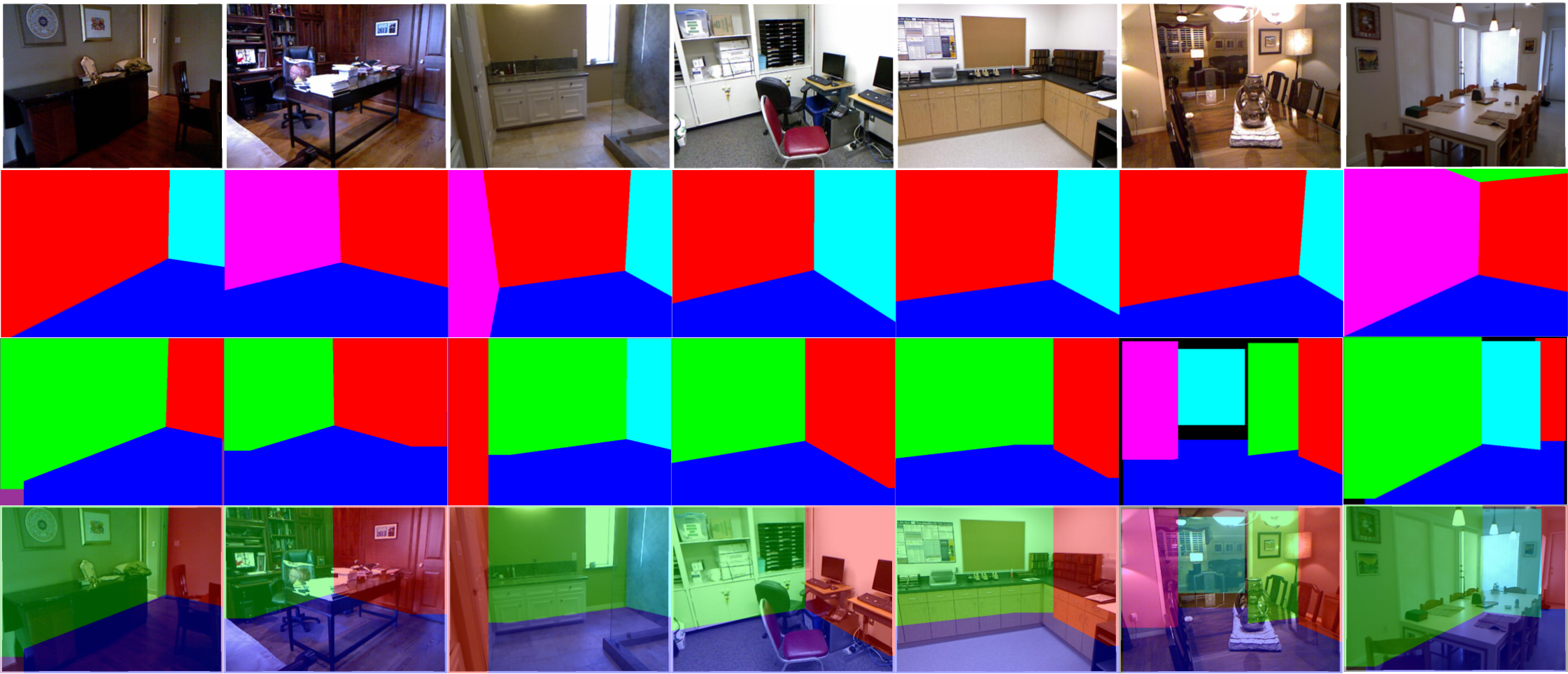}
	\caption{Example 2D room layouts produced by our method on the NYUDv2 303 dataset. The first row shows the input image. The second and third rows show the ground truth and predicted room layout segmentations, respectively. The fourth row shows the prediction overlaid onto the input image. It is worth noting that in the last two columns the rooms in the input are not box-shaped. We believe that our prediction better reflects the room shape in these cases than the ground truth.}
	\label{fig:layouts}
	\setlength{\textfloatsep}{9pt}
\end{figure*}

\subsubsection{2D Room Layout Predictions}
We demonstrate the ability for our method to capture room geometry further by producing room layout predictions for individual frames on the NYUDv2 303 dataset. Since we do not assume a box layout, we do not classify the type of wall as left, right, or front. Although floor and ceilings are obvious, these wall classes present a level of ambiguity. In an image of a room corner it is often an arbitrary choice about whether the two wall should be labelled left and front, or front and right. Therefore, to evaluate our predicted layouts we must map our wall detections to the five possible plane types in the ground truth, which we do using the Hungarian algorithm.

As can be seen with the results presented in Table \ref{tab:layouts}, we outperform all RGB methods tested, including PlaneNet \cite{liu2018planenet} and RoomNet \cite{lee2017roomnet}, designed for this task. We also believe our results could be improved by merging architectural plane detections if the plane equations are similar. This merging of detections would better reflect the ground truth box approximations used in the dataset.

Further, we present qualitative results of the 2D layouts that are produced by our method in Figure \ref{fig:layouts}. These examples help to illustrate that while the box representation of rooms is often very well suited to the room pictured, as shown in the first five columns of Figure \ref{fig:layouts}, there are cases where this approximation does not reflect the geometry of the scene particularly well. For example, in the last two columns of Figure \ref{fig:layouts}, we believe that our predictions offer a better representation of the room geometry than the ground truth.


\begin{figure*}[t]
    \centering
    \includegraphics[width=0.8\textwidth]{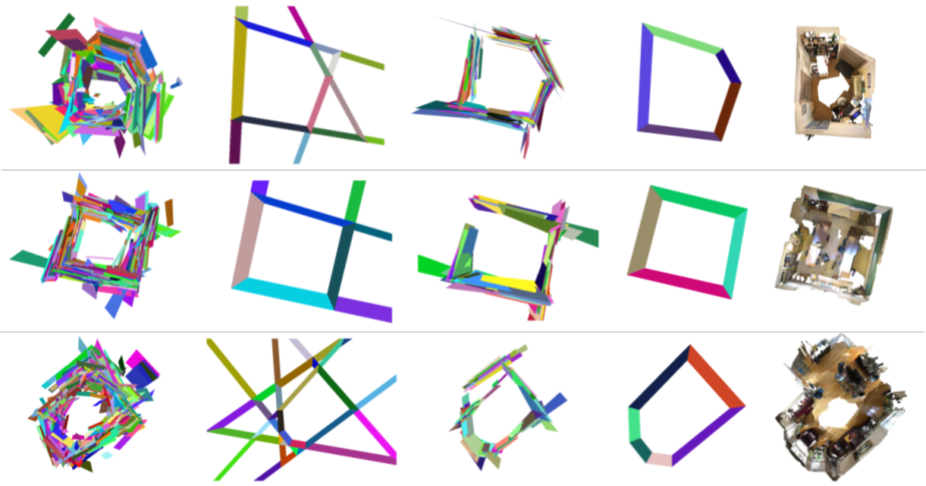}
    \caption{Each row visualises the steps of our spatial voting method for determining the room layout from Planar R-CNN measurements on a ScanNet sequence. The first column shows a subset of the raw measurements. After inferring the underlying room planes, we intersect to find the set of wall segment candidates, shown in the second column, as well as the top \textit{n} measurements from each cluster with the highest likelihood to form the group of voter patches, shown in the third column. The fourth and fifth column show the final layout estimation and the ground truth mesh, respectively.}
    \label{fig:suc_rooms}
\end{figure*}

\begin{figure*}[t]
	\centering
	\includegraphics[width=0.8\textwidth]{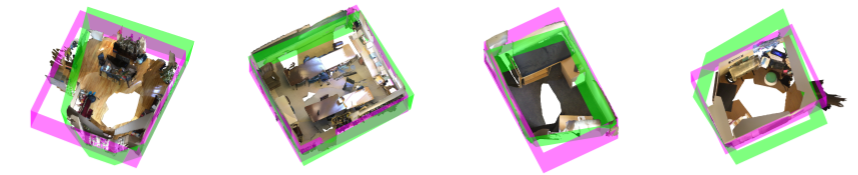}
	\caption{Manhattan boxes (magenta) fitted to the ground truth point clouds using the method provided in SUN RGB-D shown with our own unconstrained layouts (green).}
	\label{fig:box}
\end{figure*}

\subsubsection{3D Room Layout Estimation on ScanNet.}
Unfortunately, ScanNet does not provide the room layout annotations required to train our network. This meant we were unable to fine-tune or provide quantitative results on this dataset. Therefore, we provide qualitative room-layouts to demonstrate that our method intended to visualise the geometric information captured by our Planar R-CNN.

The qualitative results in Figure \ref{fig:suc_rooms} show example room layouts obtained on the ScanNet dataset. From the noisy raw plane measurements, the clustered planes encapsulate a broad representation of the room. The fourth row provides an interesting case as the upper segment of room is missed altogether. When inspecting the ground truth mesh, it was found that there are limited vertices present in the mesh. This leads us to believe that this area was not visible for much of the video. It is worth noting that, most likely because of the level of noise in the initial measurements, fine room geometry is missed in the clustering.

Although the produced room-layout estimates lack fine detail, we believe that they provide a better representation of room geometry than a Manhattan box approximation would. To demonstrate this, we produce a qualitative baseline using the Manhattan model fitting provided in SUN RGB-D. The boxes produced in Figure \ref{fig:box} were fitted to the ground truth point clouds, and thus would likely deteriorate if point clouds obtained from a monocular method were used.

We believe that the results shown in Figures \ref{fig:suc_rooms} and \ref{fig:box} demonstrate our methods ability to capture useful structure from ScanNet sequences without using depth, enforcing a Manhattan assumption, or imposing temporal consistency.

\section{Conclusions}
We have proposed a method for room layout estimation without a box or Manhattan world assumption. It combines instance segmentation implemented via an R-CNN with Gaussian Mixture Model-based probabilistic clustering, and allows for the effective reconstruction of walls which do not lie on either of the two principle aligned axes, or the vertical direction.


\subsubsection{Acknowledgements.}

We gratefully acknowledge the European Commission Project Multiple-actOrs Virtual Empathic CARegiver for the Elder (MoveCare) for financially supporting the authors for this work.

{\small
\bibliographystyle{ieee_fullname}
\bibliography{egbib}
}

\end{document}